\documentclass{article}

\usepackage{iclr2023_conference,times}
\iclrfinalcopy

\usepackage[utf8]{inputenc} %
\usepackage[T1]{fontenc}    %
\usepackage{siunitx}
\DeclareSIUnit{\million}{~\text{M}}
\DeclareSIUnit{\kilo}{~\text{K}}
\usepackage{url}            %
\usepackage{booktabs}       %
\usepackage{amsfonts}       %
\usepackage{nicefrac}       %
\usepackage{microtype}      %
\usepackage{multirow}
\usepackage{amsmath}
\usepackage{amssymb}
\usepackage{amsthm}
\usepackage[disable]{todonotes}
\usepackage{comment}
\usepackage{graphicx,subcaption}
\usepackage[subtle]{savetrees}
\usepackage[font=normalsize,labelfont={bf}]{caption}
\usepackage{parskip}
\usepackage{xr-hyper}
\usepackage{hyperref}
\usepackage{comment}
\usepackage{wrapfig}
\usepackage{subcaption}
\hypersetup{
    colorlinks=true,
    linkcolor=black,
    filecolor=magenta,      
    urlcolor=blue,
    citecolor=black
}       %

\usepackage{graphicx}

\newcommand{\bfa}{\mathbf{a}}

\newcommand{\bfx}{\mathbf{x}}
\newcommand{\bfy}{\mathbf{y}}

\newcommand{\dobydo}[2]{\frac{\partial #1}{\partial #2}}

\newcommand{\cL}{\mathcal{L}}

\newcommand{\supt}{\ensuremath{^{(t)}}}
\newcommand{\suptm}{\ensuremath{^{(t-1)}}}
\newcommand{\sups}{\ensuremath{^{(s)}}}

\newcommand{\cO}{\ensuremath{\mathcal{O}}}

\newcommand{\bbR}{\ensuremath{\mathbb{R}}}

\newcommand{\bfw}{\ensuremath{\mathbf{w}}}
\newcommand{\bfv}{\ensuremath{\mathbf{v}}}
\newcommand{\bfvth}{\ensuremath{\mathbf{\vartheta}}}

\newcommand{\bfM}{\ensuremath{\mathbf{M}}}
\newcommand{\bfbM}{\ensuremath{\bar{\mathbf{M}}}}
\newcommand{\bfJ}{\ensuremath{\mathbf{J}}}
\newcommand{\bfbc}{\ensuremath{\bar{\mathbf{c}}}}

\newcommand{\bfat}{\ensuremath{\bfa\supt}}
\newcommand{\bfatm}{\ensuremath{\bfa\suptm}}
\newcommand{\bfMt}{\ensuremath{\bfM\supt}}
\newcommand{\bfMtm}{\ensuremath{\bfM\suptm}}
\newcommand{\bfJt}{\ensuremath{\bfJ\supt}}
\newcommand{\bfbct}{\ensuremath{\bfbc\supt}}

\newcommand{\tia}{\ensuremath{\tilde{\alpha}}}
\newcommand{\tib}{\ensuremath{\tilde{\beta}}}
\newcommand{\tio}{\ensuremath{\tilde{\omega}}}

\makeatletter
\newcommand*{\addFileDependency}[1]{%
  \typeout{(#1)}
  \@addtofilelist{#1}
  \IfFileExists{#1}{}{\typeout{No file #1.}}
}
\makeatother

\title{Efficient Real Time Recurrent Learning through combined activity and parameter sparsity}

\author{Anand Subramoney$^{1,2,\footnotemark[1]{}}$\\
$^1$ Institute for Neural Computation, Ruhr University Bochum, Germany \\
$^2$ Royal Holloway, University of London \\
\texttt{anand.subramoney@rhul.ac.uk} \\
}

\begin{document}

\maketitle
\footnotetext[1]{Work done while at Ruhr University Bochum}

\begin{abstract}

Backpropagation through time (BPTT) is the standard algorithm for training recurrent neural networks (RNNs), which requires separate simulation phases for the forward and backward passes for inference and learning, respectively.
Moreover, BPTT requires storing the complete history of network states between phases, with memory consumption growing proportional to the input sequence length.
This makes BPTT unsuited for online learning and presents a challenge for implementation on low-resource real-time systems.
Real-Time Recurrent Learning (RTRL) allows online learning, and the growth of required memory is independent of sequence length.
However, RTRL suffers from exceptionally high computational costs that grow proportional to the fourth power of the state size, making RTRL computationally intractable for all but the smallest of networks.
In this work, we show that recurrent networks exhibiting high activity sparsity can reduce the computational cost of RTRL.
Moreover, combining activity and parameter sparsity can lead to significant enough savings in computational and memory costs to make RTRL practical.
Unlike previous work, this improvement in the efficiency of RTRL can be achieved without using any approximations for the learning process.

\end{abstract}

\section{Introduction}
\label{sec:introduction}

Recurrent neural networks are powerful models with a wide variety of applications ranging from language modelling~\citep{merity_Regularizing_2017} to reinforcement learning~\citep{espeholt_IMPALA_2018}.
Transformers have supplanted RNNs in many areas of task performance but are impractical for training or inference in resource-constrained environments due to their computational and memory requirements.
RNNs may provide a viable alternative, but the most commonly used training algorithm, backpropagation through time (BPTT), requires memory that grows with the length of sequences and does not allow for online updates.
Real-time recurrent learning (RTRL) allows for online updates. Memory requirements are independent of sequence length, but the computational requirements are extremely high.
This makes it impractical for larger networks -- even for a network with 100 units, each step would require on the order of $10^6$ computations.

Event-based neural networks (EvNNs) with activity sparsity have been previously proposed to make RNNs more efficient and include spiking neural networks~\citep{maass1997networks} and the EGRU~\citep{subramoney2022egru} among others.
For the EGRU, which is a form of EvNN based on GRU dynamics, \citet{subramoney2022egru} show that an EvNN with activity sparsity requires fewer operations not only for inference but also during training with BPTT.
However, due to the disadvantages of BPTT listed above, we explore using RTRL for these EvNNs.
We show that for EvNNs with activity sparsity, the computational requirements of RTRL can be reduced to be proportional to $\tib^2 n^2 p$ (rather than $n^2 p$ for dense RTRL) where $\beta = 1 - \tib$ is the sparsity of the derivative of the activations of the units in the backward pass.
The memory requirement is also reduced by a factor of $\tib$.
Moreover, we show that benefits due to activity sparsity compose well with those with parameter sparsity.
With the combination of the two sparsities, the computational cost is ameliorated significantly to be proportional to $\tio^2 \tib^2 n^2 p$ (rather than $n^2 p$ for dense RTRL) where $\omega = 1 - \tio$ is the parameter sparsity, $n$ is the size of the network state, and $p$ is the number of parameters ($p=n^2$ for a fully connected vanilla RNN).
Since $\tib$ and $\tio$ are value between 0 and 1, the savings in compute cost can be significant. 
\citet{subramoney2022egru} demonstrate that the EGRU achieves about 50\% backward sparsity in most of their experiments. 
With just activity sparsity, this translates to a multiplier of $\tib^2 = (1 - 0.5)^2 = 0.25$ or 25\% of the operations required for dense RTRL without activity sparsity.
Combined with parameter sparsity of ~80\%, the multiplier is $0.2^2 0.5^2 = 0.01$, which translates to 1\% of operations required for dense RTRL.
This is achieved without the need for approximating the RTRL updates.

The benefits of these computational savings can be realised to the fullest extent on devices with support for unstructured sparsity.
Support for unstructured sparsity has been increasing both on established hardware types such as GPUs~\citep{nvidia-2-4} and newer hardware designed for deep learning~\citep{jia2019dissecting,cerebras} as well as neuromorphic devices~\citep{davies2018loihi,hoppner2017dynamic}.

\section{Related work}
There is a long line of work exploring methods to make RTRL more efficient -- see \citet{marschall2020unified} for an extensive review.
But most previous approaches have involved finding good approximations for calculating the influence matrix in RTRL.
We use the canonical, exact equations for calculating the RTRL influence matrix while taking advantage of the natural computational advantage that arises from having an activity-sparse network such as the EvNN/EGRU.
\citet{menick2020practicala} have explored the influence of parameter sparsity in making RTRL more efficient and conclude that by itself, the gains are limited due to the form of interactions between parameters and activations in recurrent networks.
They propose approximations combined with parameter sparsity that can lead to computationally more efficient RTRL, whereas we use the exact form of RTRL updates to achieve computational efficiency.

\section{Real Time Recurrent Learning}

\begin{table}[hbtp]
\centering
\begin{tabular}{l l c c}
  \toprule
  \multicolumn{2}{l}{Method} & memory & time per step \\
  \midrule
  \multirow{2}{*}{Fully dense} & BPTT & $Tn + p$ & $n^2 + p$ \\
                               & RTRL & $n + np$ & $n^2 + n^2p$ \\
  \midrule
  \multirow{2}{*}{Sparse RTRL} & with parameter sparsity & $n + \tio np$ & $\tio n^2+\tio^2 n^2 p$ \\
                               & with activity sparsity & $\tia n + \tib np $ & $\tia n^2 + \tib^2 n^2 p$\\
                               & with both & $\tia n + \tio \tib n p$ & $\tio \tia n^2 + \tio^2 \tib^2 n^2 p$\\
  \midrule
  \multirow{2}{*}{Approximate RTRL} & SnAp-1~\citep{menick2020practicala} & $n+\tio p$ & $\tio n^2+\tio p$\\
                                    & SnAp-2~\citep{menick2020practicala} & $n+\tio^2 n p$ & $\tio n^2 + \tio^3 n^2 p$\\
\bottomrule
\end{tabular}
\caption{
    Computational and memory costs for sparse and dense methods, including approximations. 
    $T$ refers to sequence length, $n$ the number of hidden units, $p$ the number of dense parameters. $\omega = 1 - \tio$ is the level of parameter sparsity, $\alpha = 1 - \tia$ is the level of activity sparsity in the forward pass, and $\beta = 1 - \tib$ is the level of sparsity in the derivative of activation (see text for further explanation). 
    The first term of each compute cost is for the forward pass, and the second term is for updating the influence matrix.
}
\label{tab:cost-comparison}
\end{table}

Following the notation in~\citep{marschall2020unified}, we define a recurrent network that has a state $\bfat \in \bbR^n$ at each timestep $t$.
This state is updated with function $F: \bbR^m \rightarrow \bbR^n$ parameterised by the flattened recurrent parameter vector $\bfw \in \bbR^p$, and input parameter vector $\bfw_\text{in} \in \bbR^{n_\text{in}}$.
Here $m = n + n_{\text{in}} + 1$ is the total number of input dimensions, $n$ being the number of hidden units, $n_\text{in}$ the dimensionality of the input, and the $+1$ to include a bias term.
For a densely connected vanilla RNN $p=n^2$. 
We first consider activity sparsity and derive the computational savings for a fully connected vanilla RNN (the derivation is equivalent for architectures such as the GRU or EGRU).

Consider a network whose dynamics is defined by
\begin{equation}
    \bfat = F\left(\bfatm,\; \bfx^{(t)}; \bfw \right).
\end{equation}
An output $\bfy^{(t)} \in \bbR^{n_\text{out}}$ is computed at every timestep using a readout function $\bfy^{(t)} = F_\text{out}\left(\bfat; \bfw_o\right)$, where $\bfw_o$ are the readout parameters.
This output determines the instantaneous loss at every time step, calculated as $L\supt = L\left(\bfy\supt,\; \bfy_\text{target}\supt\right)$, where $L$ is some loss function, say MSE or cross-entropy loss.
The goal of training is to minimize the total loss over all timesteps $\cL = \sum_t L\supt$ using gradient descent on all the parameters.

For updating the recurrent parameters, we need to calculate $\dobydo{\cL}{\bfw}$.
We consider how this quantity can be calculated using RTRL, starting with
\begin{equation}
    \dobydo{\cL}{\bfw} = \sum_{t=0}^{T} \dobydo{L\supt}{\bfw} = \sum_t \sum_{s=0}^{t} \dobydo{L\supt}{\bfw\sups},
\end{equation}
where $\bfw\sups$ refers to the application of the parameters $\bfw$ at timestep $s$ (equivalent to the partial derivative of $L\supt$).
Using chain rule, this can be written as 
\begin{equation}
    \dobydo{L\supt}{\bfw} = \left(\dobydo{\bfat}{\bfw}\right)^T \dobydo{L\supt}{\bfat} \equiv \left(\bfMt\right)^T \bfbct,
\end{equation}
where $.^T$ refers to matrix transpose.
We refer to $\bfM\supt = \partial \bfat / \partial \bfw \in \bbR^{n \times p}$ as the influence matrix, and $\bfbc\supt = \partial L\supt / \partial \bfat \in \bbR^n$ as the credit assignment vector.

In RTRL, \bfMt~is defined by the following recursive relationship:
\begin{equation}\label{eq:main-recursive}
    \bfMt \equiv \bfJ\supt \bfM\suptm + \bfbM\supt,
\end{equation}
where the jacobian $\bfJ\supt = \partial \bfat / \partial \bfatm \in \bbR^{n \times n}$ and the immediate influence matrix $\bfbM\supt = \partial\bfat/\partial\bfw\supt \in \bbR^{n \times p}$ are used.

The main computational cost of RTRL arises from having to multiply \bfJt~and \bfMtm, which are $n \times n$ and $n \times p$ matrices respectively, which requires $\cO(n^2 p)$ computations. 
Moreover, we need to store $\bfMt$ at every step, requiring $\cO(n p)$ memory.
For a fully connected vanilla RNN, $p = n^2$, requiring $\cO(n^4)$ computations and $\cO(n^3)$ memory.

\section{Efficient RTRL with activity sparsity}

\begin{figure}
    \begin{minipage}[c]{0.39\textwidth}
        \centering
        \includegraphics[width=0.7\textwidth]{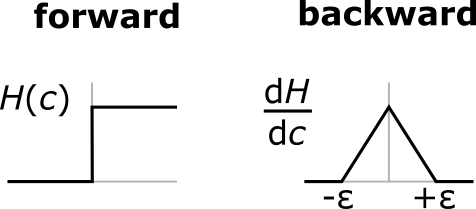}
        \caption{Illustration of surrogate gradient (pseudo-derivative)\label{fig:pseudo-derivative}}
    \end{minipage}
    ~
    \begin{minipage}[c]{0.59\textwidth}
        \centering
        \includegraphics[width=\textwidth]{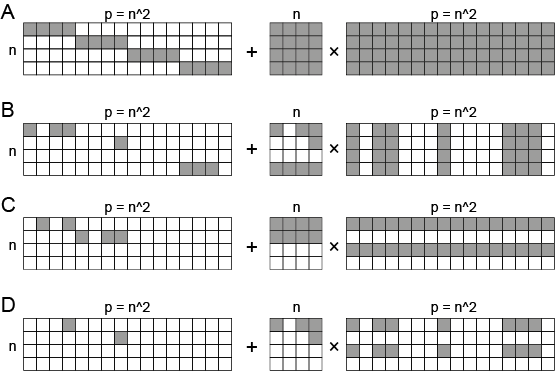}
        \caption{Illustration of RTRL updates with various types of sparsity. 
            (A) Dense network.
            (B) With only parameter sparsity.
            (C) With only activity sparsity.
            (D) With both activity and parameter sparsity.
        \label{fig:matrix}}
    \end{minipage}
    \vspace{-10pt}
\end{figure}

We now show how activity sparsity can ameliorate these computational and memory requirements.
Any network with an activation function and derivative that induces sparsity, and with the appropriate dynamics can potentially benefit from  computational and memory savings.
For simplicity, we focus on showing this for a network based on the EGRU~\citep{subramoney2022egru} that uses a Heaviside step function with a pseudo-derivative.
Specifically, consider a network where the internal state is gated by a threshold function
\begin{equation}
    \bfat = H\left(\bfv\supt\right),
\end{equation}
where $H$ is the Heaviside step function, and $\bfv\supt = F\left(\bfatm, \bfx\supt\right) - \bfvth$, $\bfvth$ defined as a vector containing `thresholds' for each unit of the network.
As in \citet{subramoney2022egru}, the gradient of the Heaviside step function is defined using a surrogate function (or pseudo-derivative) to be
$H'(\bfv_k\supt) = \gamma \text{max}\{0, 1 - \frac{|\bfv_k\supt|}{2\epsilon}\}$, where $\gamma$ defines the height of the pseudo-derivative, and $\epsilon$ defines the width of the pseudo-derivative. 
See Fig.~\ref{fig:pseudo-derivative} for an illustration.

At timestep $t$, let $\alpha\supt$ be the activity-sparsity of the network i.e. $\alpha\supt$ fraction of the units have zero activation.
For a unit $k$ with zero activation, $\bfat_k = 0$ because $\bfv\supt_k \leq 0$.
Let $\beta\supt$ be the fraction of the units which have a zero derivative.
For a unit $k$ with zero derivative, $H'(\bfv\supt_k) = 0$ because $\bfv\supt_k  > \epsilon \text{ or } \bfv\supt_k < -\epsilon$.

Now considering a single element of the Jacobian, 
\begin{align}\label{eq:jacobian}
    \bfJ_{kl}\supt = \dobydo{\bfa_k\supt}{\bfa_l\suptm} = H'\left(\bfv_k\supt\right) \dobydo{\bfv_k\supt}{\bfa_l\suptm},
\end{align}
it is clear that this term is zero for all units $k$ where $\bfv\supt_k  > \epsilon \text{ or } \bfv\supt_k < -\epsilon$.
This implies that entire rows  of the matrix \bfJ~are zero, when the unit state $\bfv_k\supt$ lies in that range ($\beta\supt \times n$ rows to be precise).

Similarly, for the immediate influence matrix $\bfbM\supt$,
\begin{equation}\label{eq:barM}
    \bfbM\supt_{kp} = \dobydo{\bfat_k}{\bfw\supt_p} = H'\left(\bfv\supt_k\right) \dobydo{\bfv_k\supt}{\bfw\supt_p},
\end{equation}
all rows corresponding to units whose states $\bfv_k\supt$ lie between $\pm \epsilon$ are zero ($\beta\supt \times n$ rows to be precise).
This is in addition to the default sparsity of $\bfbM$ since only terms immediately local to each unit are non-zero.

Each recursive update of the influence matrix can be written as
\begin{align}
    \bfM_{kp}\supt &= \sum_l \bfJ_{kl}\supt \bfM_{lp}\suptm + \bfbM_{kp}\supt, \\
                   &= \sum_l H'\left(\bfv_k\supt\right) \dobydo{\bfv_k\supt}{\bfa_l\suptm} \bfM_{lp}\suptm + H'\left(\bfv\supt_k\right) \dobydo{\bfv_k\supt}{\bfw\supt_p} \\
                   &= H'\left(\bfv\supt_k\right) \left[\sum_l \dobydo{\bfv_k\supt}{\bfa_l\suptm} \bfM_{lp}\suptm + \dobydo{\bfv_k\supt}{\bfw\supt_p} \right].
\end{align}
This implies that at each timestep $t$, $\beta\supt \times n$ rows of the matrix $\bfM\supt$ are fully $0$.

Taken together with Eqns.~\eqref{eq:jacobian},~\eqref{eq:barM}, at each time $t$, $\bfJ\supt$ has $\beta\supt$ fraction of rows zero, and $\bfM\suptm$ has $\beta\suptm$ fraction of rows zero. 
Defining $\tib\supt = 1 - \beta\supt$ and $\tib\suptm = 1 - \beta\suptm$, the matrix multiplication is on the order of $\cO(\tib\supt n \times \tib\suptm n \times p) = \cO(\tib\suptm \tib\supt n^2 p)$.
And at each timestep $t$, we would require $\cO(\tib\supt n \times p) = \cO(\tib\supt n  p)$ storage.

\section{Efficient RTRL with activity AND parameter sparsity}

As described in \citet{menick2020practicala}, for recurrent networks in the form of a vanilla RNN or GRU (and by extension EGRU), parameter sparsity introduces sparsity in elements of $\bfJ$, and sets entire columns of $\bfbM$ to zero. 
If the parameter sparsity pattern is fixed, then corresponding columns of $\bfM$ remain zero across timesteps  as well.
Therefore, with a parameter sparsity of $\omega = 1 - \tio$, and activity sparsity, the compute requirement for the influence matrix updates from Eq.~\eqref{eq:main-recursive} is $\cO(\tib\supt n \times \tio\tib\suptm n \times \tio p) = \cO(\tib\supt \tib\suptm \tio^2 n^2 p)$.
The memory requirement for storing the influence matrix is $\cO(\tib\supt n \times \tio p) = \cO(\tib\supt \tio n p)$.

Fig.~\ref{fig:matrix} demonstrates various different cases without and with activity and parameter sparsity. 
The sparsity of the influence matrix due to activity sparsity depends on the sparsity of the derivative of the activation function. And in this case, corresponding rows of $\bfJ$ and $\bfM$ and $\bfbM$ become zeros when the derivative of the activation function becomes zero for a neuron.
In the case of parameter sparsity, elements of $\bfJ$ and columns of $\bfM$ and $\bfbM$ become zero when a parameter is zero.
See Table~\ref{tab:cost-comparison} for the full comparison.

\section{Results}

\begin{figure}
    \centering
    \includegraphics[width=\textwidth]{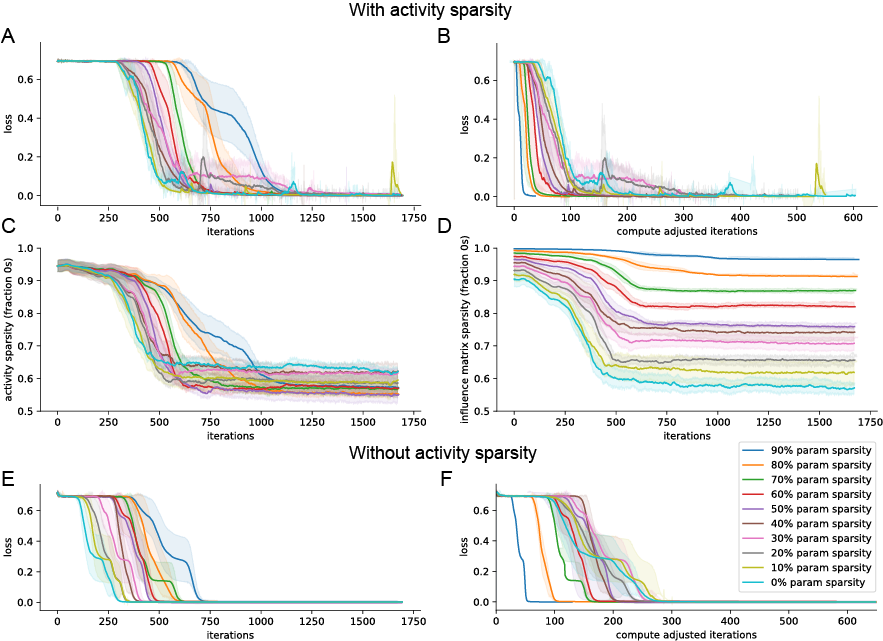}
    \caption{Computational savings and sparsity results on synthetic dataset. Mean and std err over 5 runs. \label{fig:results}}
    \vspace{-10pt}
\end{figure}

We use RTRL with combined activity and parameter sparsity on a synthetic classification task where a two-dimensional spiral unwinding over time is classified as clockwise or anti-clockwise.
The dataset consisted of 10,000 randomly generated spirals of 17 timesteps length assigned to one of the two classes depending on the orientation of the spiral.
We trained an EGRU~\citep{subramoney2022egru} with 16 hidden units for 1700 iterations with Adam~\citep{Adam2015} and a batch size of 32.
Note that most variants converge well before 1700 iterations.
For cases where the parameter sparsity was non-zero, we chose a fixed random sparsity mask at initialisation and trained the network with this sparsity mask throughout.

The learning curves of a network with activity sparsity and various levels of parameter sparsity are shown in Fig.~\ref{fig:results}A.
Those for the case without activity sparsity are shown in Fig.~\ref{fig:results}E.
The x axis here shows the number of iterations/epochs corresponding to parameter updates.
We consider a measure of total compute with the compute adjusted iteration, defined as the cumulative sum of the computational savings factor $\tio^2 \tib^2$ (or $\tio^2$).
Note that the compute adjusted iteration is an analytical measure for the total compute used in an optimal case where the underlying hardware architecture is optimised for the algorithm, and does not take memory access and  computational bottlenecks into consideration.
Using this measure, we can see in Figs.~\ref{fig:results}B and F that the combination of high (90\%) parameter sparsity with activity sparsity converges with the least total compute.
Figs.~\ref{fig:results}C and D show the activity and influence matrix sparsity respectively for different degrees of parameter sparsity for a network with activity sparsity enabled.
Since the parameter sparsity is fixed at initialisation does not change,
the influence matrix sparsity also remains fixed throughout training when activity sparsity is turned off.

\section{Discussion}\label{sec:discussion}

We demonstrate the advantage of combining activity and parameter sparsity both analytically and empirically on a small synthetic dataset.
We demonstrate the result on an event-based network architecture (EGRU~\citep{subramoney2022egru}) that is naturally activity sparse.
Even with modest levels of activity and parameter sparsity, the computational savings realisable in RTRL is quite significant, making RTRL potentially applicable to real world tasks.
While we use fixed parameter sparsity, methods such as~\citep{bellec2018deep} allow for optimising the parameter sparsity pattern during training.

In order to fully realise the compute savings, an optimised implementation that runs on hardware supporting unstructured sparsity is required.
With the two forms of sparsity considered here, a message-passing implementation of RTRL can provide a new practical event-based way of training recurrent neural networks on neuromorphic devices as well.
On the whole, given the appropriate hardware substrate, RTRL with a combination of activity and parameter sparsity can provide a practical and competitive alternative to BPTT.

\section{Acknowledgements}
We acknowledge the use of Fenix Infrastructure resources, which are partially funded from the European Union's Horizon 2020 research and innovation programme through the ICEI project under the grant agreement No. 800858. 
AS was funded by the Ministry of Culture and Science of the State of North Rhine-Westphalia, Germany during this work.
AS would like to thank Mark Sch\"one, Khaleelulla Khan Nazeer and David Kappel for helpful discussions and comments on this manuscript, and Laurenz Wiskott for institutional support.

\bibliography{references}

\begin{thebibliography}{13}
\providecommand{\natexlab}[1]{#1}
\providecommand{\url}[1]{\texttt{#1}}
\expandafter\ifx\csname urlstyle\endcsname\relax
  \providecommand{\doi}[1]{doi: #1}\else
  \providecommand{\doi}{doi: \begingroup \urlstyle{rm}\Url}\fi

\bibitem[Bellec et~al.(2018)Bellec, Kappel, Maass, and
  Legenstein]{bellec2018deep}
G.~Bellec, D.~Kappel, W.~Maass, and R.~Legenstein.
\newblock Deep {{Rewiring}}: {{Training}} very sparse deep networks.
\newblock In \emph{International {{Conference}} on {{Learning
  Representations}}}, Feb. 2018.
\newblock URL \url{https://openreview.net/forum?id=BJ_wN01C-}.

\bibitem[Davies et~al.(2018)Davies, Srinivasa, Lin, Chinya, Cao, Choday, Dimou,
  Joshi, Imam, Jain, et~al.]{davies2018loihi}
M.~Davies, N.~Srinivasa, T.-H. Lin, G.~Chinya, Y.~Cao, S.~H. Choday, G.~Dimou,
  P.~Joshi, N.~Imam, S.~Jain, et~al.
\newblock Loihi: A neuromorphic manycore processor with on-chip learning.
\newblock \emph{Ieee Micro}, 38\penalty0 (1):\penalty0 82--99, 2018.

\bibitem[Espeholt et~al.(2018)Espeholt, Soyer, Munos, Simonyan, Mnih, Ward,
  Doron, Firoiu, Harley, Dunning, Legg, and Kavukcuoglu]{espeholt_IMPALA_2018}
L.~Espeholt, H.~Soyer, R.~Munos, K.~Simonyan, V.~Mnih, T.~Ward, Y.~Doron,
  V.~Firoiu, T.~Harley, I.~Dunning, S.~Legg, and K.~Kavukcuoglu.
\newblock {{IMPALA}}: {{Scalable Distributed Deep-RL}} with {{Importance
  Weighted Actor-Learner Architectures}}.
\newblock \emph{arXiv:1802.01561 [cs]}, Feb. 2018.
\newblock URL \url{http://arxiv.org/abs/1802.01561}.

\bibitem[H{\"o}ppner et~al.(2017)H{\"o}ppner, Yan, Vogginger, Dixius, Partzsch,
  Neum{\"a}rker, Hartmann, Schiefer, Scholze, Ellguth,
  et~al.]{hoppner2017dynamic}
S.~H{\"o}ppner, Y.~Yan, B.~Vogginger, A.~Dixius, J.~Partzsch, F.~Neum{\"a}rker,
  S.~Hartmann, S.~Schiefer, S.~Scholze, G.~Ellguth, et~al.
\newblock Dynamic voltage and frequency scaling for neuromorphic many-core
  systems.
\newblock In \emph{2017 IEEE International Symposium on Circuits and Systems
  (ISCAS)}, pages 1--4. IEEE, 2017.

\bibitem[Jia et~al.(2019)Jia, Tillman, Maggioni, and
  Scarpazza]{jia2019dissecting}
Z.~Jia, B.~Tillman, M.~Maggioni, and D.~P. Scarpazza.
\newblock Dissecting the graphcore ipu architecture via microbenchmarking.
\newblock \emph{arXiv preprint arXiv:1912.03413}, 2019.

\bibitem[Kingma and Ba(2015)]{Adam2015}
D.~P. Kingma and J.~Ba.
\newblock Adam: {A} method for stochastic optimization.
\newblock In Y.~Bengio and Y.~LeCun, editors, \emph{3rd International
  Conference on Learning Representations, {ICLR} 2015, San Diego, CA, USA, May
  7-9, 2015, Conference Track Proceedings}, 2015.
\newblock URL \url{http://arxiv.org/abs/1412.6980}.

\bibitem[Lewington(2021)]{cerebras}
R.~Lewington.
\newblock Cerebras systems: Achieving industry best ai performance through a
  systems approach.
\newblock
  \url{https://8968533.fs1.hubspotusercontent-na1.net/hubfs/8968533/Whitepapers/Cerebras-CS-2-Whitepaper.pdf},
  2021.
\newblock [Online; accessed 07-Feb-2023].

\bibitem[Maass(1997)]{maass1997networks}
W.~Maass.
\newblock Networks of spiking neurons: the third generation of neural network
  models.
\newblock \emph{Neural networks}, 10\penalty0 (9):\penalty0 1659--1671, 1997.

\bibitem[Marschall et~al.(2020)Marschall, Cho, and Savin]{marschall2020unified}
O.~Marschall, K.~Cho, and C.~Savin.
\newblock A {{Unified Framework}} of {{Online Learning Algorithms}} for
  {{Training Recurrent Neural Networks}}.
\newblock \emph{Journal of Machine Learning Research}, 21\penalty0
  (135):\penalty0 1--34, 2020.
\newblock ISSN 1533-7928.
\newblock URL \url{http://jmlr.org/papers/v21/19-562.html}.

\bibitem[Menick et~al.(2020)Menick, Elsen, Evci, Osindero, Simonyan, and
  Graves]{menick2020practicala}
J.~Menick, E.~Elsen, U.~Evci, S.~Osindero, K.~Simonyan, and A.~Graves.
\newblock Practical {{Real Time Recurrent Learning}} with a {{Sparse
  Approximation}}.
\newblock In \emph{International {{Conference}} on {{Learning
  Representations}}}, Sept. 2020.
\newblock URL \url{https://openreview.net/forum?id=q3KSThy2GwB}.

\bibitem[Merity et~al.(2017)Merity, Keskar, and
  Socher]{merity_Regularizing_2017}
S.~Merity, N.~S. Keskar, and R.~Socher.
\newblock Regularizing and {{Optimizing LSTM Language Models}}.
\newblock \emph{arXiv:1708.02182 [cs]}, Aug. 2017.

\bibitem[Pool et~al.(2021)Pool, Sawarkar, and Rodge]{nvidia-2-4}
J.~Pool, A.~Sawarkar, and J.~Rodge.
\newblock {Accelerating Inference with Sparsity Using the NVIDIA Ampere
  Architecture and NVIDIA TensorRT}.
\newblock
  \url{https://developer.nvidia.com/blog/accelerating-inference-with-sparsity-using-ampere-and-tensorrt/},
  2021.
\newblock [Online; accessed 07-Feb-2023].

\bibitem[Subramoney et~al.(2022)Subramoney, Nazeer, Sch{\"o}ne, Mayr, and
  Kappel]{subramoney2022egru}
A.~Subramoney, K.~K. Nazeer, M.~Sch{\"o}ne, C.~Mayr, and D.~Kappel.
\newblock {{EGRU}}: {{Event-based GRU}} for activity-sparse inference and
  learning, June 2022.
\newblock URL \url{http://arxiv.org/abs/2206.06178}.

\end{thebibliography}

\end{document}